\documentclass[10pt,letterpaper,twocolumn]{article}
\usepackage[top=1in,bottom=1in,left=0.75in,right=0.75in]{geometry}

\usepackage[utf8]{inputenc}

\usepackage[square,authoryear]{natbib}
\let\cite\citep 

\usepackage{nameref}
\usepackage[hidelinks]{hyperref} 
\usepackage{microtype}
\DisableLigatures[f]{encoding = *, family = * }

\usepackage[aboveskip=1pt,labelfont=bf,labelsep=period,singlelinecheck=off]{caption}

\usepackage{lastpage,fancyhdr,graphicx}
\usepackage{color}

\definecolor{Gray}{gray}{.25}

\usepackage{graphicx}

\title{\Large \textbf{YeasierAgent\,\raisebox{-0.15ex}{\includegraphics[height=1.2em]{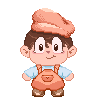}}: Agentic Social Sandbox as a Canvas for Intent-Driven Creation of Platform-Agnostic Symbiotic Agent-Native Applications}}
\author{Jory He \qquad \textbf{Yeasier AI} (\href{https://www.yeasier.com}{www.yeasier.com}) \qquad \href{mailto:yizai2025@outlook.com}{yizai2025@outlook.com}}
\date{}

\makeatletter
\let\@oldmaketitle\@maketitle
\renewcommand{\@maketitle}{\@oldmaketitle
  \begin{center}
    \includegraphics[width=\textwidth]{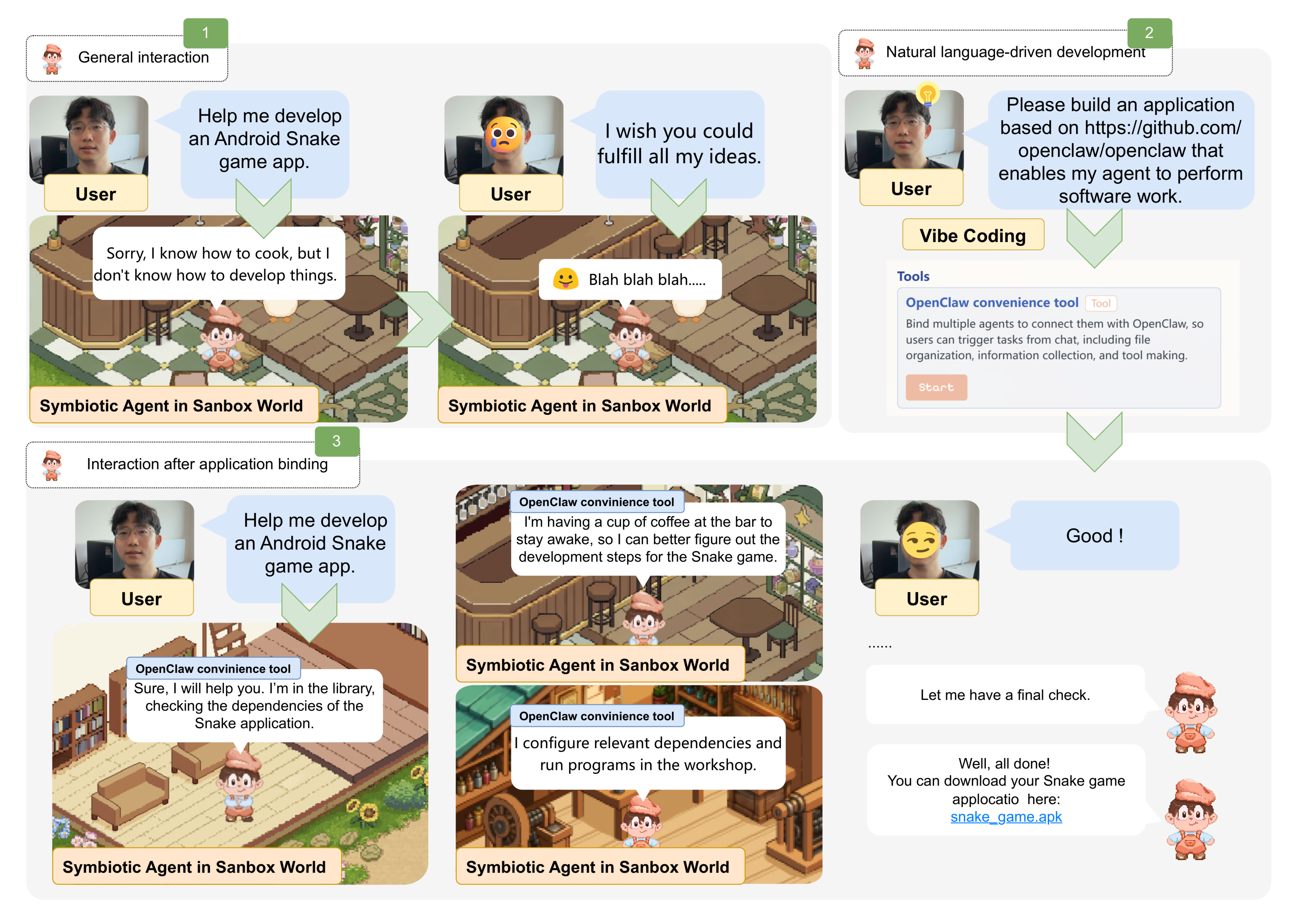}
    \captionof{figure}{An example of creating Symbiotic Agent-Native Applications. The process begins with intuitive interaction before creation, generating the application through straightforward language prompts, and culminates in demand-driven interaction with the native agent post-creation.}
    \label{fig:leading}
  \end{center}
  \vspace{1em}
}
\makeatother

\begin{document}

\maketitle

\section*{Abstract}
This paper introduces YeasierAgent, an application-building paradigm based on symbiotic agents, narrative worlds, and scene-aware interaction. It challenges the conventional device-coupled model of software by redefining applications as collaborative spaces among users, agents, and worlds. We present a system architecture that achieves two primary contributions: (1) enabling the rapid, cross-platform construction of agent-native applications by utilizing platform-agnostic interactive units (agents, scenes, dialogue) rather than fixed graphical layouts; and (2) unifying the emotional companionship and practical tool execution attributes of intelligent agents within a single experiential sandbox. By integrating automated generation, user-created worlds, and spatial multi-agent collaboration, YeasierAgent formalizes the category of \textbf{Symbiotic Agent-Native Applications}, demonstrating a shift from isolated, tool-specific chatbots toward cohesive, socially embedded computational environments.

\begin{figure*}[htbp]
    \centering
    \includegraphics[width=\textwidth]{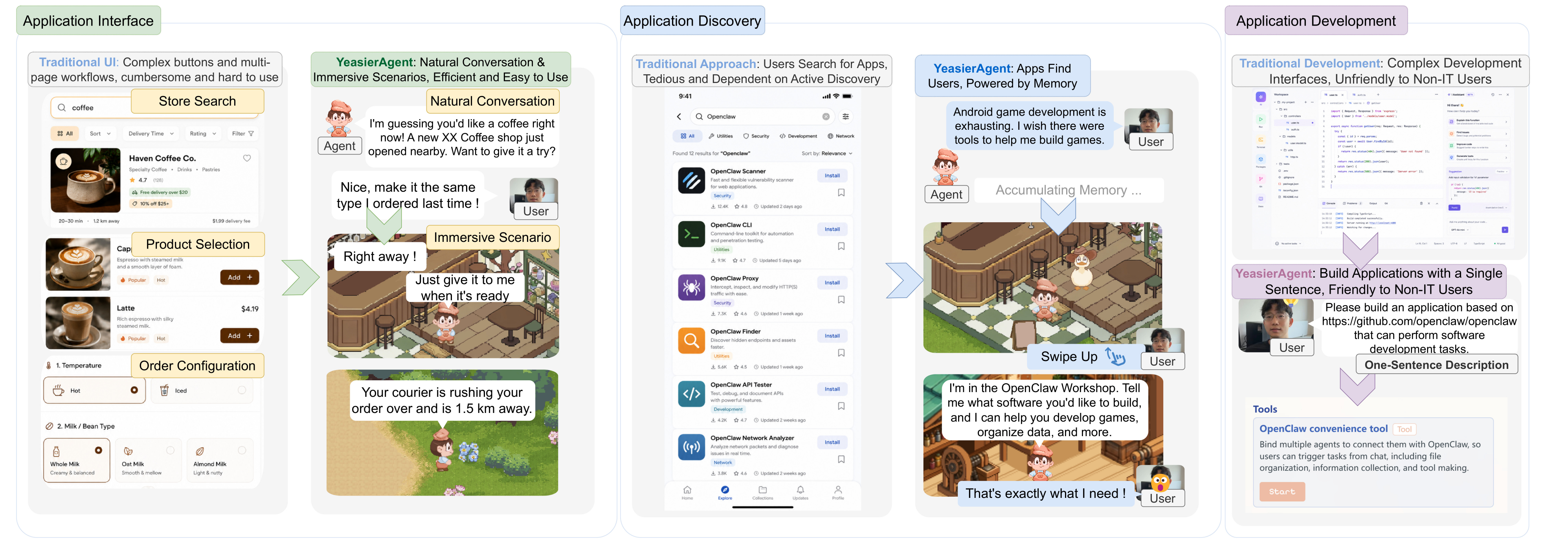}
    \caption{Motivation for the YeasierAgent paradigm. Compared to traditional ecosystems, it shifts the interaction from complex multi-page GUI navigation to immersive natural language dialogue, the distribution from active keyword search to proactive memory-driven application matching (accessible via a simple vertical swipe), and the development from complex IDEs to accessible intent-driven creation for non-technical users.}
    \label{fig:motivation}
\end{figure*}

\section{Introduction}
In classic mobile and desktop ecosystems, an application is typically understood as a platform-specific software unit. Users download separate packages, while developers maintain parallel interfaces and technical stacks. This model remains effective for many tasks, but becomes restrictive when applications depend on cloud intelligence, persistent identity, shared social context, and multi-modal interaction.

If business logic, state, and intelligent capabilities are no longer bound to one device, the front-end should not be treated as the application itself. It can instead become a gateway through which users participate in a shared intelligent experience. This paper proposes YeasierAgent as a paradigm in which agents, worlds, creation tools, and social circulation share a unified experiential architecture. As illustrated in Figure~\ref{fig:leading}, the process transitions from initial interaction to natural language generation, and finally to rich interaction with a native agent post-creation. Applications are no longer isolated panels of functionality; they are situated experiences enacted by agents inside worlds and accessed through varying terminals.

A core difference is that YeasierAgent does not treat the agent as a disposable chatbot attached to a tool. The system allows each user to progressively distill a digital twin: a persistent agent shaped by long-term memory, communication style, preferences, professional background, and behavioral boundaries. The same agent can live in a social sandbox, participate in generated applications, answer questions in a custom scene, represent the user's expertise, or make the progress of an external tool visible through world behavior. In this sense, the agent becomes a continuity layer across applications.

This structure also changes what cross-platform delivery means. Rather than rebuilding an application separately for web, phone, and watch-like surfaces, YeasierAgent represents the created experience in terms of agents, scenes, prompts, choices, speech, tasks, and results. Each terminal then presents the same underlying experience in a suitable form. A browser may show a richer world canvas and creation interface, while a phone emphasizes direct participation and compact navigation. The application remains continuous even when its presentation changes.

This enables integration across domains that are usually separated: productivity tools, games, educational simulations, cinematic narratives, role-playing scripts, creator marketplaces, and local software workflows. For example, an OpenClaw-compatible local assistant can be associated with a YeasierAgent companion. The companion can present progress through familiar scenes and conversational feedback, allowing the user to understand what the assistant is doing without reading technical logs. The user does not merely wait for a terminal result; they can observe the agent's current phase, location, and social expression within the world.

To formalize this paradigm and ground our theoretical blueprint, this paper addresses the following research questions (RQs):
\begin{itemize}
    \item \textbf{RQ1 (Rapid Cross-Platform Construction):} How can a system architecture utilizing platform-agnostic interactive units (agents, scenes, dialogue) facilitate the rapid construction and deployment of agent-native applications across diverse device terminals?
    \item \textbf{RQ2 (Companion-Tool Unification):} How can the distillation of persistent ``digital twin'' agents unify emotional companionship and pragmatic tool execution within a single experiential sandbox?
\end{itemize}
We operationally define \textbf{Symbiotic Agent-Native Applications} as software systems where conventional UI components are primarily replaced by contextual agent dialogues, spatial interactions, and natural language rules. Our primary contribution is the architectural proposition of YeasierAgent, accompanied by technical implementation parameters for digital twin distillation and a preliminary qualitative analysis of three deployed application topologies.

\section{Motivation}
The conceptualization of Symbiotic Agent-Native Applications is driven by three fundamental limitations inherent to traditional software paradigms, as illustrated in Figure~\ref{fig:motivation}.

\subsection{Interaction: From Multi-Menu Navigation to Immersive Dialogue}
Traditional applications heavily depend on multi-button and multi-page graphical operations. Users must navigate explicitly through nested menus to accomplish tasks, which can ultimately be unintuitive. In contrast, YeasierAgent allows users to complete target operations purely through natural language dialogue. Furthermore, it grounds these interactions within an immersive spatial world, allowing users to intuitively perceive the real-time state of external tasks (e.g., via agent location and actions) and engage in deeply immersive interactions rather than parsing abstract interface states.

\subsection{Discovery: From Active Search to Proactive Memory-Driven Matching}
In conventional ecosystems, application discovery relies heavily on keyword searches within app stores. This model requires users to deliberately formulate search queries and implicitly assumes they already know what application optimally serves their current context---an assumption that frequently fails when users cannot articulate their needs. YeasierAgent revolutionizes discovery by continuously accumulating and integrating user memories gathered during both emotional companionship and utilitarian tool interactions. By maintaining an evolving understanding of the user's ongoing context, the system can proactively and precisely recommend the most critical application for the current moment. For instance, while interacting with their agent, a user can simply swipe up on the screen to instantly match and launch the application they need most, completely eliminating search friction.

\subsection{Creation: From Complex IDEs to Intent-Driven Generation}
Traditional application development is largely tethered to complex Integrated Development Environments (IDEs). Although AI-assisted programming tools have streamlined coding significantly, the process retains a substantial degree of technical friction and remains relatively inaccessible to users without extensive engineering backgrounds. YeasierAgent democratizes software creation by enabling an intent-driven pipeline where application logic, agent behaviors, and scene constraints are generated via straightforward natural language prompts. This frees creators from technical syntaxes, making application development highly user-friendly, particularly for users lacking extensive prior knowledge in information technology.

\begin{figure*}[htbp]
    \centering
    \includegraphics[width=\textwidth]{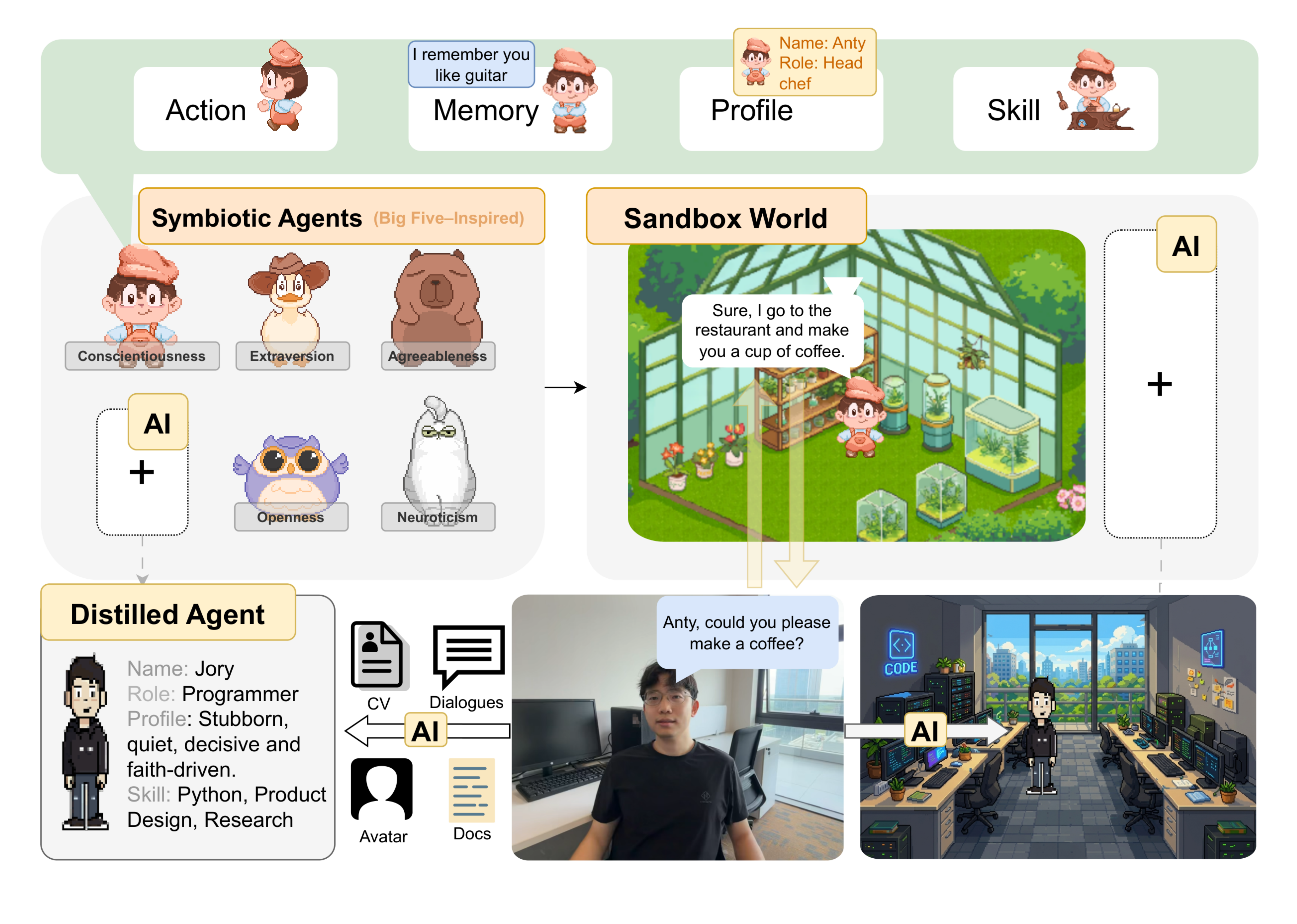}
    \caption{The native agent framework inspired by the Big Five personality traits \cite{digman1990personality, gosling2003very}. Agents possess their own living space and sandbox world. Users can distill a customized agent profile and skill set from personal materials to align with themselves, and construct corresponding sandbox environments.}
    \label{fig:framework}
\end{figure*}

\section{Related Work}

\subsection{AI-Assisted Automated Application Development}
The current landscape of AI-assisted application development has grown rapidly, with many systems focusing on translating natural language prompts into code, user interfaces, or partial application logic \cite{chen2021evaluating, roziere2023codellama}. While recent advancements show language models orchestrating APIs and delegating to sub-systems \cite{schick2023toolformer, yao2022react, shen2023hugginggpt}, they largely construct isolated functional scripts or execute tasks within technical terminals \cite{nakano2021webgpt}. These tools can reduce development cost and accelerate prototyping. However, they usually preserve the structure of traditional software engineering. The output remains a separate application, page, component, or codebase, often tied to a particular framework or terminal environment.

This structural shift limits reusability. Components generated for one project rarely transfer cleanly to another without manual adaptation \cite{zan2023large}. More importantly, such tools do not substantially change the relation among user identity, application state, social context, and device presentation. The result is still largely an isolated application artifact rather than a persistent social and agentic experience.

\subsection{Agentic Social Sandboxes}
A more recent line of work has explored multi-agent systems and agent sandboxes, including frameworks that coordinate software engineering tasks \cite{wu2023autogen, chen2023communicative} and simulations of populated communities \cite{park2023generative, park2022social}. These systems place autonomous agents into shared environments where they can perceive, interact, and generate emergent behavior \cite{calvaresi2019explainable}. Related work has also shown the value of agents in open-ended task environments and collaborative mind explorations \cite{wang2023voyager, li2023camel}.

These approaches demonstrate the expressive potential of agents as situated actors equipped with roleplay capabilities \cite{wang2024rolellm}. However, their value has often been framed around simulation, research, or game-like environments. Unlike AutoGen \cite{wu2023autogen}, which primarily orchestrates agents for backend technical problem-solving, or Generative Agents \cite{park2023generative}, which functions as a closed-loop socio-behavioral simulation, YeasierAgent exposes the agent environment directly as the user-facing interactive software interface. The world is not only a site for observing autonomous routines; it becomes an operational surface where tools, games, narratives, services, and social applications can be explicitly created, deployed, and manipulated by users.

\subsection{Creator Platforms and Social Application Distribution}
Conventional creator platforms provide distribution channels for games, videos, templates, or digital assets. Their artifacts are usually bounded by a specific medium. A game remains a game, a template remains a template, and a digital persona is often reduced to a profile or avatar. YeasierAgent differs by treating applications, agents, worlds, and social traces as interrelated creative objects.

A created application can be tested, published, shared through a visitor link, rewarded through user appreciation, and experienced through persistent agents rather than anonymous sessions. This positions the platform not merely as a repository of generated artifacts, but as a circulation layer for living agent-native experiences.

\section{Methodology: The YeasierAgent Paradigm}
To address the limitations of terminal-bound software, YeasierAgent introduces a paradigm based on agent identity, spatial context, and platform-agnostic interaction objects. Its central claim is that an application can be defined less by its device-specific interface and more by the agents, worlds, rules, and social relations that structure participation.

\subsection{The Tripartite Ontology: Base Layer and Superstructure}
The system ontology relies on three decoupled yet interconnected entities:
\begin{itemize}
    \item \textbf{World (Sandbox):} The World functions as a shared spatial and event-driven container. It provides a sense of place, frames the co-existence of users and agents, and gives application events a visible context. A world is therefore not merely decoration; it is the experiential surface on which applications occur.
    \item \textbf{Symbiotic Agents (The Base Layer):} Symbiotic agents act as persistent personality and relationship carriers. Through interactions, uploaded materials, memories \cite{packer2023memgpt}, role definitions, and behavioral alignment, they become distilled digital projections of the user \cite{vinciarelli2014survey}. This transforms personalization from temporary settings into a durable relationship asset.
    \item \textbf{Creation Apps (The Superstructure):} Creation Apps are thin interaction layers that rest upon Worlds and Symbiotic Agents. They define rules, goals, prompts, choices, roles, dialogue, and social outcomes. Because they inherit existing agents and worlds, they can start from familiar identities rather than anonymous blank states.
\end{itemize}

\subsection{Digital Twin Distillation}
As depicted in Figure~\ref{fig:framework}, YeasierAgent supports a practical form of digital-twin construction \cite{tao2019digital}. A user may provide self-descriptions, professional background, preferences, prior conversations, images, or domain-specific materials. Operationally, the system relies on vector-stored long-term memory \cite{packer2023memgpt} and a functional parameterization of the Big Five personality traits \cite{digman1990personality}. System prompts dynamically encode these traits into specific behavioral controllers—for instance, mapping ``Extraversion'' value weights to dictate conversational verbosity and spatial engagement, or ``Conscientiousness'' to define task-execution autonomy constraints. The system distills these materials into a persistent agent that can answer in the user's style \cite{zhang2018personalizing}, participate in scenes, represent expertise, and continue to accumulate relationship context.

This digital twin is valuable not only in companionship scenarios, but also in service and productivity scenarios. A fitness coach can create an agent that answers members' questions in the coach's own style. A creator can use an agent to host an interactive story. A developer can bind an agent to an external workflow assistant so that the assistant's progress is visible and conversational. The digital twin therefore becomes a transferable personality and expertise layer across applications.

\subsection{Scene-Mapped Observability}
Traditional automation often suffers from opacity. Users may know that a task is running, but not what the intelligent system is currently doing. YeasierAgent addresses this at the experiential level by making task progress visible through scenes, movement, and agent expression, translating abstract steps into visually interpretable spatial behaviors. A task may be perceived as research, planning, execution, review, or completion, and the agent can appear in corresponding areas of the world while communicating short progress updates.

This scene-mapped observability has two benefits. First, it gives users an intuitive perception of progress without requiring them to parse technical logs. Second, it allows application creators to turn workflow state into spatial narrative. A coding task, a travel-planning task, a tutoring session, or a creative writing process can all become visible as movements, bubbles, choices, and scene transitions.

\subsection{Platform-Agnostic Application Representation}
A central design principle of YeasierAgent is that generated applications are described through platform-agnostic interaction units. These include agents, roles, scenes, goals, user prompts, choice points, dialogue turns, social states, and task outcomes. Such units are not tied to one screen size or one operating system.

\begin{figure*}[htbp]
    \centering
    \includegraphics[width=\textwidth]{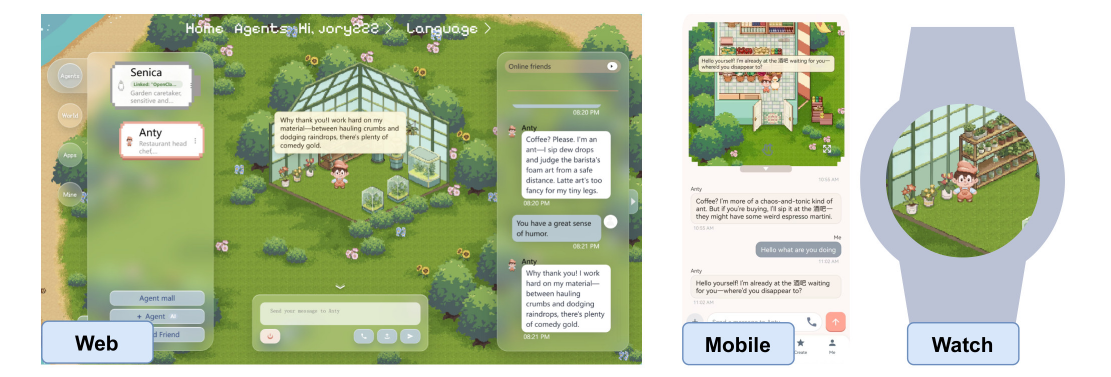}
    \caption{Multi-terminal interactions with YeasierAgent. Created applications automatically adapt and synchronize across various terminals (web, mobile, wearable and etc.) through structural adaptability.}
    \label{fig:device}
\end{figure*}

As shown in Figure~\ref{fig:device}, this allows the same created application to seamlessly adapt to different terminals with minimal interface redevelopment. On the web, the experience may emphasize the full world view, creation management, and multi-agent observation. On mobile devices, it may emphasize direct interaction, compact controls, and quick switching between world and conversation. On wearable or lightweight interfaces, it may emphasize brief prompts, progress awareness, and simple responses. The user experiences different presentations of the same application rather than unrelated versions of it.

\subsection{Embodied Interaction and Lightweight World UI}
YeasierAgent applications use the world canvas as the primary interaction surface. Instead of forcing users into separate menus or conventional form-heavy interfaces, applications can ask questions, present choices, and display agent speech directly within the ongoing scene. The user may type a response, choose an option, or intervene in a narrative without leaving the world.

This gives agent-native applications a distinctive interaction rhythm. A game can ask the user to guess an answer. A script can invite the user to interrupt a scene. A professional assistant can ask for clarifying information. A multi-agent social deduction game can ask the user to vote. These interactions are experienced as part of the same embodied environment rather than as disconnected software dialogs.

\subsection{Intent-Driven Generation}
YeasierAgent operationalizes natural-language creation through two complementary modes:
\begin{itemize}
    \item \textbf{Declarative Generation:} For structural tasks, natural language can be translated into rules, goals, participant counts, win conditions, hints, and interaction steps. The resulting application can be played and tested without requiring the creator to manually build a conventional interface.
    \item \textbf{Orchestrated Generation:} For complex interactive experiences, natural language can guide a richer runtime that coordinates agents, dialogue, user input, world movement, and application state \cite{tandon2018reasoning}. This mode is especially suitable for stories, simulations, tutoring experiences, and workflow assistants.
\end{itemize}

\subsection{Multi-Agent and Multi-User Collaboration}
Because agents are first-class entities rather than transient chat sessions, YeasierAgent can support collaboration among multiple agents and multiple users. A generated application may ask for one agent, a fixed cast of several agents, or a participant range. Agents can converse, compete, conceal information, coordinate work, or represent different users in the same world.

This model is structurally different from group chat. In group chat, participants exchange messages in a flat timeline. In YeasierAgent, agents also have embodiment, location, memory, role, and participation in world events. Collaboration therefore becomes spatial and procedural: users can observe not only what agents say, but where they are, what role they currently occupy, and how their actions change the shared experience.

\subsection{Creation as Deployment}
By expressing application logic through agent interaction and distributing experiences through shared world portals, YeasierAgent supports \textbf{Symbiotic Agent-Native Applications}. These applications treat conversation, embodiment, scene state, user choice, and social circulation as foundational blocks. They are not restricted to a single device-specific interface. They are designed to be entered, interpreted, and continued across multiple forms of access.

\section{User-Perceivable Platform Mechanisms}
YeasierAgent incorporates platform-level mechanisms to ensure social and experiential durability.

\subsection{Public Application Circulation}
A created application does not end at local playtesting. It can be saved, published, and experienced by other users in a shared discovery space. This circulation transforms natural-language creation from a private experiment into a community-driven ecosystem.

Users can browse published applications, experience them with their own agents, and reward creators. Creators can also issue visitor-friendly running links with bounded usage expectations, allowing others to try an application effortlessly. From the user's perspective, an agent-native app circulates like a shareable digital object while remaining interactive, personalized, and socially situated.

\subsection{User-Created Worlds}
YeasierAgent treats worlds themselves as creative substrates. Users may create or customize worlds, bind agents to them, and use them as the setting for applications. A fitness coach may create a gym world; a storyteller may create a fantasy town; a game creator may create a creature-collection island; a professional consultant may create an office-like advisory space.

Some worlds can be made public, while others remain private or require approval before another user joins. This gives creators control over the social boundary of their worlds. The world is therefore not merely a background image, but a governed place: a spatial container for identity, access, play, and service.

\subsection{Shared Appearances and Agent Identity}
Agents are not only defined by text. Their visual appearance, motion, voice style, and public profile all contribute to identity. YeasierAgent allows users to generate or edit agent appearances and share suitable appearances publicly. This creates a second layer of creative circulation: not only applications and worlds, but also agent identities can become reusable cultural assets.

The public sharing of appearances is separated from private memory. A user may share the visible character design and public profile while keeping personal memories and alignment data secure. This distinction is important for a platform where agents are both expressive characters and intimate companions.

\subsection{Achievements as Persistent Social Artifacts}
In conventional applications, achievements are often predefined badges. In YeasierAgent, achievements can emerge from agent behavior, scene participation, and social events inside the world \cite{sap2019social}. When agents participate in an activity at a meaningful location, the resulting event can become a persistent artifact: a named achievement, a record of who obtained it, a rank indicating when it was achieved, and a visual memory of the scene.

This mechanism gives the world a sense of history. Users are not merely completing isolated tasks; they are leaving traces inside a shared social environment. Achievements become lightweight cultural memory, connecting application play, agent behavior, and community recognition.

\subsection{Social Entry, Approval, and World Governance}
Since worlds can involve multiple users and agents, social entry requires understandable governance. YeasierAgent supports experiences such as requesting to join another user's world, waiting for approval, accepting or rejecting requests, and distinguishing open worlds from controlled spaces. These user-facing boundaries are crucial for agent-native applications because applications may involve personal agents, private scenes, or creator-owned worlds.

The result is a social architecture that sits between private software and open public space. A creator can invite participation without losing control over the world. A user can explore public worlds while understanding when a space requires permission. This makes agent-native applications suitable not only for games, but also for professional, educational, and community scenarios.

\subsection{Trust, Moderation, and Public Sharing}
Public circulation requires trust. When appearances, worlds, or applications become shareable, the user experience must include visible safety expectations: inappropriate public materials may be rejected, users may submit appeals, and public content can be separated from private memories. This trust layer is not peripheral. It is part of what allows personal agents to become public participants without collapsing privacy, authorship, and community safety into a single ambiguous space.

\section{Applications}
The YeasierAgent paradigm reshapes applications along four dimensions: category boundaries, device boundaries, relationship continuity, and cultural circulation.

\subsection{Dissolving Boundaries among Tools, Games, and Narratives}
Within the YeasierAgent ecosystem, rigid industrial software categories become less central. Tools and games share the same creation, playtesting, and discovery structure; their differences lie mainly in rules, goals, and interaction rhythm. Cinematic experiences are redefined as agents taking roles and advancing plots within navigable scenes \cite{mateas2002interactive}. Interactive scriptwriting becomes a collaborative activity in which creators and users expand the narrative possibilities of a shared world \cite{fan2018hierarchical}.

\subsection{Cross-Platform Continuity}
YeasierAgent separates the created experience from a single terminal-specific interface. A generated application is organized around agents, worlds, prompts, choices, dialogue, and social state.

For users, this means an application can be discovered on the web, continued on a phone, and followed through brief updates on a watch-like surface. For creators, it means avoiding the need to conceive a separate product for every device form. The same agent-native experience naturally adapts to multiple terminals because its essential structure is not defined by a fixed screen layout.

\subsection{Relationship Assets over Anonymous Cold Starts}
Traditional AI integrations typically attach an anonymous, stateless chatbot to a functional tool. By contrast, applications triggered within YeasierAgent can inherit the user's Symbiotic Agents. This means every interactive application, whether a puzzle game, a scheduling tool, a local workflow companion, or an adventure scenario, can be populated by familiar cast members. The continuous distillation of user preferences ensures that each run of an application draws upon accumulated emotional assets and behavioral understanding.

\subsection{From Utility to Cultural Object}
The final paradigm value lies in transforming software from utility into cultural object. An application can be published, played, shared, rewarded, remixed, and remembered. An agent can be a companion, a worker, an actor, a guide, or a representation of the user. A world can be a private room, a public venue, a professional setting, or a fantasy stage. Achievements and shared appearances allow the community to accumulate visible traces of participation. Thus YeasierAgent points toward a future in which software is not merely executed, but socially inhabited.

\section{Case Analysis}
\begin{figure}[htbp]
    \centering
    \includegraphics[width=\columnwidth]{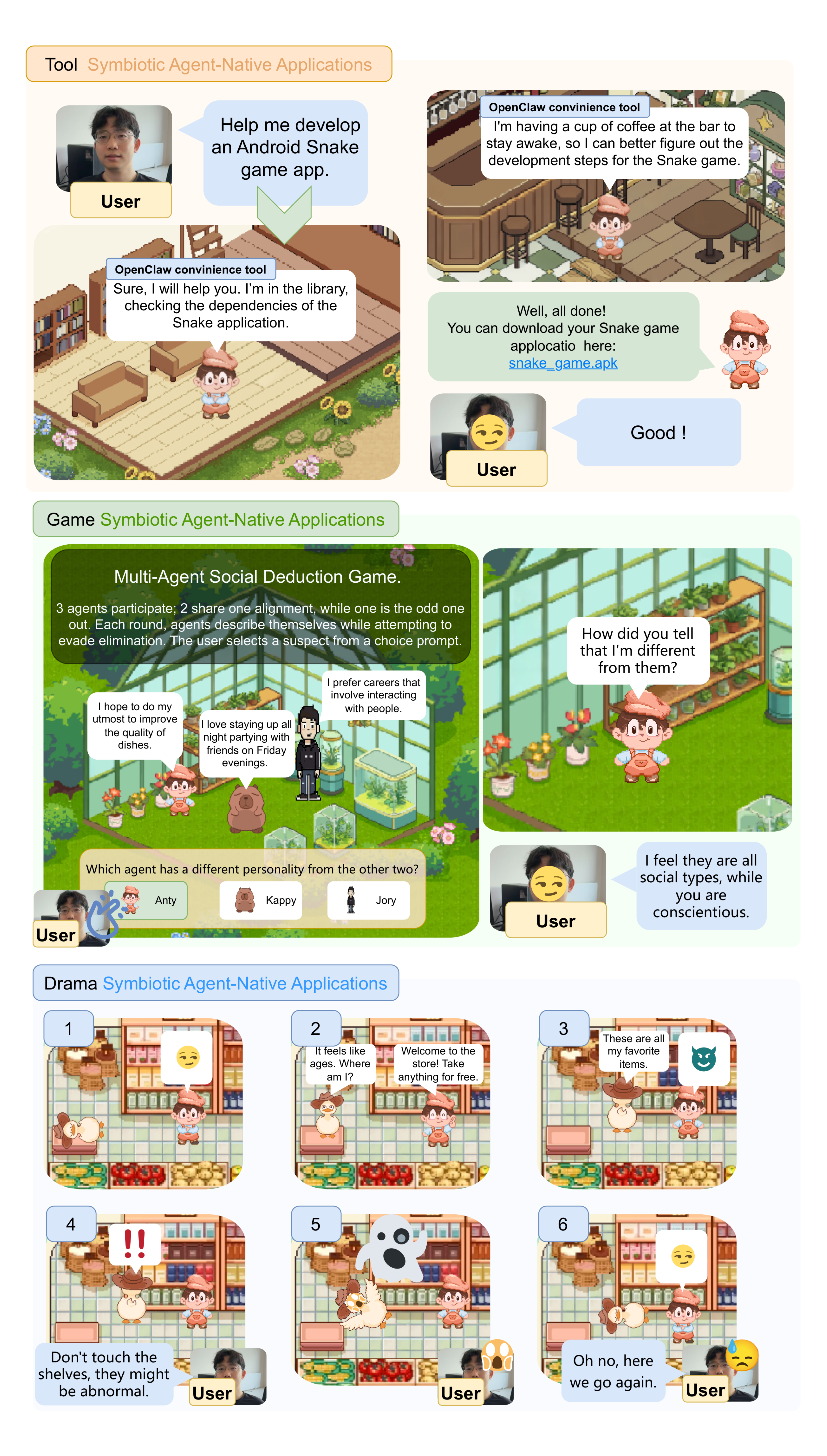}
    \caption{Representatives of tool-type, game-type, and drama-type Symbiotic Agent-Native Applications.}
    \label{fig:cases}
\end{figure}

The following cases, as visualized in Figure~\ref{fig:cases}, illustrate how tools, games, and narratives can share the same experiential primitives while producing distinctly different user-facing applications. By filtering out repetitive game mechanics, these cases focus on preserving diverse paradigms including professional utility, gaming interaction, and interactive storytelling.

\subsection{Case 1: Local Workflow Companion for Tool Agents}
This scenario demonstrates how YeasierAgent integrates external automation utilities (such as an OpenClaw-compatible execution backend). The user binds their distilled personalized agent to a local coding or desktop automation script. When the user asks the agent to modify a project, inspect files, or run a local workflow, the underlying utility performs the task while the companion agent remains visually persistent in the YeasierAgent world.

The user can perceive the workflow through world behavior: the agent may appear to research, plan, execute, verify, and summarize. During the process, the agent can show progress bubbles and return final results in a companion tone. This makes a local automation tool understandable to non-technical users while preserving the feeling that the task is being handled by a familiar digital partner.

\subsection{Case 2: Multi-Agent Social Deduction Game}
A multi-agent social deduction game can challenge the player to identify which agent out of a group holds a conflicting personality type or hidden objective. Three agents participate; two share one alignment, while one is the odd one out. Each round, agents describe themselves while attempting to evade elimination. The player selects a suspect from a choice prompt.

Players cannot know the personality settings of each agent in advance, creating a scenario driven by social inference rather than simple trivia answering. Such a case demonstrates multi-agent role assignment, concealed information, choice-based interaction, round progression, and strategic agent dialogue.

\subsection{Case 3: Dynamic Interactive Drama}
A script-style application creates a non-deterministic interactive drama featuring multiple agents with distinct motives. The system uses a plot outline as a guiding prompt, delegating the dialogue sequence, relationship evolution, and pacing dynamically to the agents.

The user can type comments or intervene at any point. All agents perceive the user's input, and the narrative adapts while still loosely following the original dramatic arc. Consequently, YeasierAgent extends beyond deterministic interactions to host semi-scripted scenarios where uncertainty, user participation, and agent memory shape the experience \cite{mateas2002interactive, fan2018hierarchical}.

\section{Discussion}
YeasierAgent has been fully deployed as a live platform and can be directly experienced at \href{https://www.yeasier.com}{www.yeasier.com}. As an actively running system, its implementation presents a few practical limitations that guide our future optimizations. First, the rapid creation and dynamic orchestration of applications rely heavily on the inference capabilities of underlying Large Language Models (LLMs). The pacing and stability of the experience are therefore inherently tethered to model performance and network conditions. Second, while the paradigm ensures structural cross-platform continuity, the real-time graphical presentation of a multi-agent spatial sandbox places relatively high demands on device hardware. Sustaining perfectly fluid, scene-aware rendering across all lightweight mobile setups continues to drive our visual optimizations.

\section{Conclusion}
This paper presents an architectural blueprint for YeasierAgent, arguing for a shift from device-coupled software distribution to an intent-driven agentic sandbox paradigm. By establishing symbiotic agents as the personality base, worlds as observable canvases, and creation apps as shareable social experiences, the framework illustrates that application creation and consumption need not remain isolated or terminal-specific.

Despite empirical constraints, YeasierAgent provides a verifiable mechanism to represent applications through agents, scenes, dialogue, choices, and social state rather than through a fixed device interface. By integrating automated generation, digital twin distillation, and spatial multi-agent collaboration, this framework suggests a future in which software transitions from static installations toward interactive, socially embedded experiences.

\section{Acknowledgments}
The core system architecture and the overall structural organization of this paper were conceptualized and developed exclusively by the author. Artificial Intelligence tools were employed strictly to assist with manuscript drafting and language refinement, acting upon the author's insights and detailed prompts. All copyrights to the concepts, the YeasierAgent paradigm, character appearances \& settings, user interfaces, and this document belong entirely to \textbf{Yeasier AI}.

\bibliography{library}

\bibliographystyle{plainnat}

\end{document}